\def\year{2019}\relax
\documentclass[letterpaper]{article} 
\usepackage{aaai19}  
\usepackage{times}  
\usepackage{helvet}  
\usepackage{courier}  
\usepackage{url,multirow}  
\usepackage{graphicx,booktabs}  
\usepackage{amsmath}
\usepackage{amsfonts}

\begin{document}
%
\title{Knowledge-Grounded Response Generation with \\ Deep Attentional Latent-Variable Model}
\author{Hao-Tong Ye\quad Kai-Ling Lo\quad Shang-Yu Su\quad Yun-Nung Chen\\
 National Taiwan University, Taiwan \\
 \texttt{\{b04902102,b04902010,f05921117\}@ntu.edu.tw\quad y.v.chen@ieee.org}
 }
\maketitle
\begin{abstract}
End-to-end dialogue generation has achieved promising results without using handcrafted features and attributes specific for each task and corpus. However, one of the fatal drawbacks in such approaches is that they are unable to generate informative utterances, so it limits their usage from some real-world conversational applications. This paper attempts at generating diverse and informative responses with a variational generation model, which contains a joint attention mechanism conditioning on the information from both dialogue contexts and extra knowledge.
\end{abstract}

\section{Introduction}
Dialogue-related research can be mainly categorized into two branches: (1) task-oriented: the system trying to help users complete a certain task (2) chit-chat: the system that can handle casual conversations that do not belong to any specific domain.
Recently, how to bridge these two branches has become a new research direction in conversation modeling, where the system can generate useful and fact-grounded  responses via external knowledge without domain constraints  \cite{hori2017end,DSTC7,ghazvininejad2017knowledge}.

Prior work showed that end-to-end neural models are capable of generating sound responses for chit-chat dialogues in a data-driven way, without using handcrafted features specific for each corpus or different task~\cite{vinyals2015neural,sordoni2015neural,li2016deep,gao2018neural}. 
However, such systems still highly rely on the information stored in training corpora, which is constrained by time, space, and speakers during data collection.
Also, the systems lack the direct access to external information and the knowledge-grounded mechanism; therefore they cannot effectively retrieve real-world common senses and facts in order to respond properly.
This fundamental limitation makes the end-to-end systems difficult to complete tasks \cite{li2017end,peng2018adversarial} or generate fact-grounded chit-chat \cite{ghazvininejad2017knowledge}.

On the other hand, for traditional dialogue systems, we can easily insert external knowledge and facts into the model with the cost of detailed hand-coding features, which requires a large amount of pre-processing and data labeling.
For those tasks or corpora related to complex information or professional knowledge, pre-processing and annotations are difficult to acquire, thus making this approach impractical.

In this work, we propose an end-to-end variational model with the attention mechanism that models the interactions between dialogue contexts and external knowledge.
This model is capable of balancing between scalability and generalization of neural models and provides more factual and knowledge-grounded responses compared to the traditional systems. 
Such extension is especially important for a conversational model deployed in systems requiring more relevant and informative interactions (e.g. the recommendation system).

To test the ability of generating knowledge-grounded responses, the seventh Dialog System Technology Challenge (DSTC7) proposed a benchmark Reddit dataset, in which the conversations are accompanied with a link to an external webpage that may contain related facts and knowledge.
A dataset example is shown in Table \ref{tab:dataset_example}, where the last two responses share the same contexts, and the fact retrieved by our model contains related knowledge given the conversation.

\begin{table*}[t!]
\begin{center}
\begin{tabular}{p{17cm}}
\toprule
\textbf{Conversation:} \\
til monty python member terry gilliam was author j . k rowling 's first choice to direct the first harry potter movie , but was rejected for chris columbus . inan interview he said " i was the perfect guy to do harry potter ... i mean , chris columbus ' versions are terrible . just dull . pedestrian " https://en.wikipedia.org/wiki/terry\_gilliam \\
---- gilliam would have been great - but we 'd still be waiting on the second movie . \\
-------- he should do an animated version \\
------------ harry potter \& the giant soft gradient foot \\
---- they hired chris columbus due to his experience directing child actors . \\
-------- i also think he 's really good at seeing things from a kid 's imagination . those first 2 movies really seemed \\ 
\hspace{0.9cm} like someone went into my head and said " ok we 're going to film a movie here!" \\
-------- came here to say this. iirc, he was hired specifically because he was good with kids... which gilliam had \\
\hspace{0.9cm} little experience with. i think they turned out very well, very true to the books. \\
\midrule
\textbf{Retrieved top-1 fact:} \\
j . k . rowling , the author of the harry potter series , is a fan of gilliam's work . consequently , he was rowling's first choice to direct harry potter and the philosopher's stone in 2000 , but warner bros . ultimately chose chris columbus for the job . [ 32 ] in response to this decision , gilliam said that " i was the perfect guy to do harry potter . i remember leaving the meeting , getting in my car , and driving for about two hours along mulholland drive just so angry . i mean , chris columbus ' versions are terrible . just dull . pedestrian . " [ 33 ] in 2006 , gilliam said that he found alfonso cuarón ' s harry potter and the prisoner of azkaban to be " really good ... much closer to what i would've done . " [ 34 ] in retrospect , however , gilliam has stated that he wouldn't have liked to direct any potter film . in a 2005 interview with total film , he said that he would not enjoy working on such an expensive project because of interference from studio executives . [ 35 ] \\
\bottomrule
\end{tabular}
\end{center}
\vspace{-2mm}
\caption{Dataset example from subreddit \textit{todayilearned}. The horizontal lines indicate the tree-structure of the conversation; we can see that the last two responses share the same contexts. The shown fact retrieved by our model is considered the most relevant to the given conversation among all facts extracted by the official script from the wikipedia page.}
\label{tab:dataset_example}
\end{table*}

\section{Proposed Approach}
The task is to generate a suitable response that contains grounded knowledge or factual information given its conversation contexts.

\subsection{Model Framework}
The main difference between this task and others is the context-relevant facts, which are retrieved from the website links mentioned at the beginning of conversations.
This external knowledge provides our model cues about how to ground the information in the response.
Therefore, we first build a retrieval model to effectively obtain the fact containing the relevant knowledge, and then learn the conversation model to generate the knowledge-grounded response.
Below we describe the detail of the proposed conversation model, where given a conversation con texts and its related facts, the goal is to generate the next probable sentence with informative knowledge.

\subsection{Conversation Model}

For each conversation, our model takes dialogue contexts $\mathbf{C}$ and context-relevant facts $\mathbf{F}$ as the input, and outputs the fact-grounded response $\mathbf{R}$.
Specifically, $\mathbf{C} = \{c_i\}_{i=1}^{N_c}$, where $c_n = \{c_{n,j}\}_{j=1}^{T^c_n}$ is a sequence of word embeddings in the $n$-th utterance of the conversation.
For the fact, $\mathbf{F} = \{f_i\}_{i=1}^{N_f}$, where $f_n = \{f_{n,j}\}_{j=1}^{T^f_n}$ is a sequence of word embeddings of $n$-th fact.
The generated response is formulated as $\mathbf{R} = \{r_i\}_{i=1}^{T^r}$.
In our model, we treat the conversation utterances and facts as two sequences, with a special token used to separate utterances in a dialogue or a fact;
that is, the contexts and facts are turned into $\mathbf{C} = \{c_i\}_{i=1}^N$ and $\mathbf{F} = \{f_j\}_{j=1}^M$ respectively, where $c_i$ and $f_j$ are word embedding vectors.

\begin{figure*}[t!]
  \centering
  \includegraphics[width=0.75\textwidth]{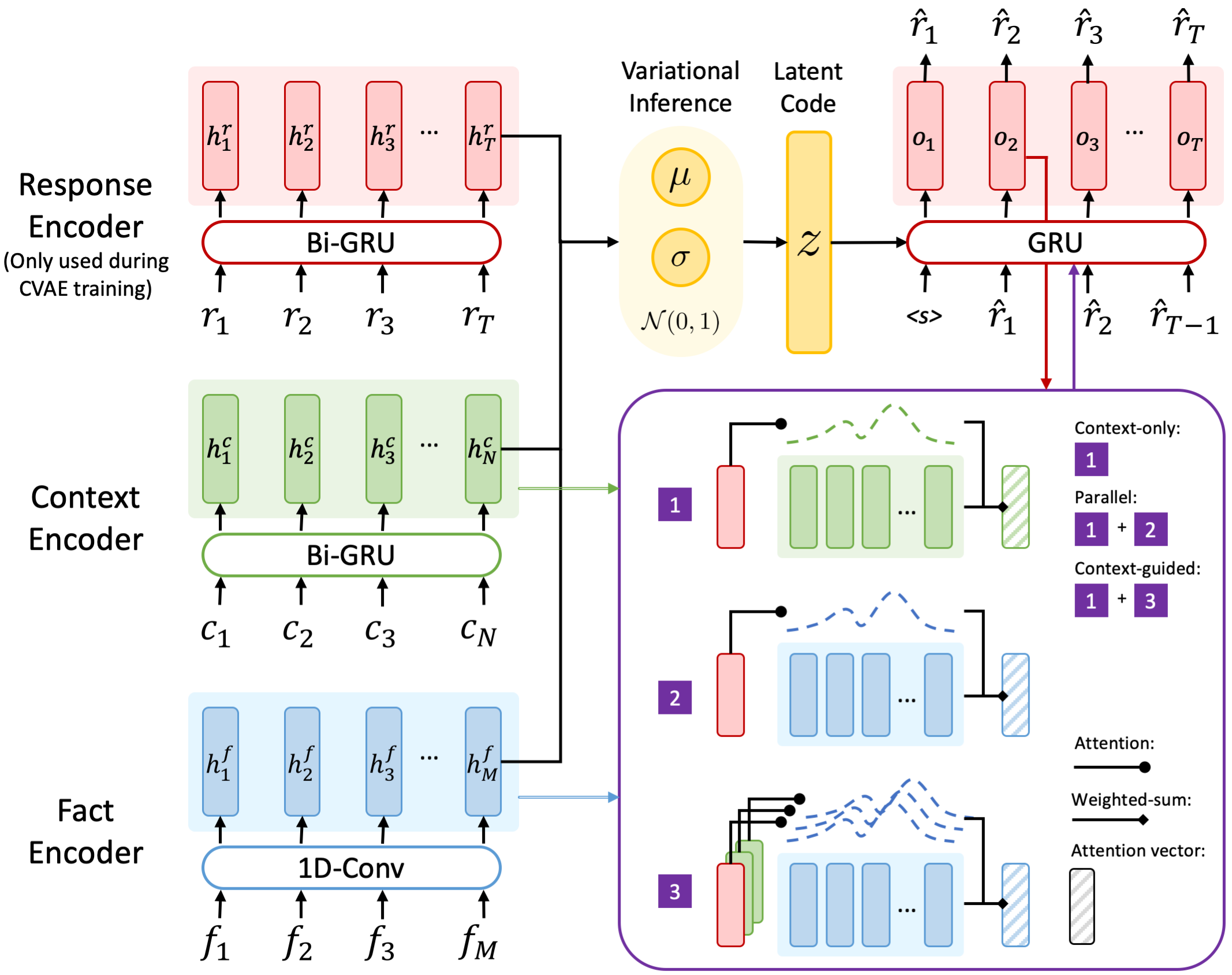}
  \caption{Illustration of the proposed model architecture.}
  \label{fig:model}
\end{figure*}

First, we use two separate encoders, $\text{Enc}_C$ and $\text{Enc}_H$, to encode the dialogue contexts and facts respectively.
The encoded contexts and facts $H_C = \{h^c_i\}_{i=1}^N$, $H_F = \{h^f_j\}_{j=1}^M$ are fed into the attention module, and then the decoder generates the fact-grounded response.
In our model, with the encoded contexts and facts, the decoder generates the response in an auto-regressive way, which is commonly called as a sequence-to-sequence model.
For each step, the output of the decoder $o_t$ is calculated from previous output $o_{t-1}$ and the encoded information $H_C$ and $H_F$:
\begin{equation}
o_t = \text{Dec}(o_{t-1}, \text{Attn}(o_{t-1}, H_C, H_F)).    
\end{equation}
The output of the decoder, $o_t$, is then projected to the vocabulary through a linear layer followed by a \texttt{softmax} activation.
The proposed model is illustrated in Figure~\ref{fig:model}, where there are several encoders that focus on different types of information.
During generation, this model proposes 1) a \emph{fact-grounded attention} mechanism that can explicitly consider the contexts snd facts and 2) a \emph{conditional variational generation} model that can produce diverse and informative responses.
The detail of two module is described below.

\subsection{Fact-Grounded Attention}
In order to capture the relations between these three types of information, dialogue contexts, facts, and responses, we apply three attention variants to model their interactions~\cite{bahdanau2014neural}: \emph{context-only attention}, \emph{parallel attention}, and \emph{context-guided fact attention} detailed below.

\subsubsection{Context-Only Attention}
One simple attention baseline only uses the information from contexts to generate the response.
That is, with the last-step output $o_{t-1}$ and the encoded information $H_C, H_F$, the attention is calculated as:
\begin{equation}
\begin{aligned}
\text{Attn}&(o_{t-1}, H_C, H_F) = \sum_{i=1}^N \alpha^t_i h^c_i, \\
e^t_i &= v_c^T \text{tanh}(W^c_1 o_{t-1} + W^c_2 h^c_i) + b_c, \\
\alpha^t &= \text{softmax}(e^t),
\end{aligned}
\end{equation}
where $v_c$, $W^c_1$, $W^c_2$, and $b_c$ are trainable parameters.

\subsubsection{Parallel Attention}
In order to well utilize the facts, one trivial solution is to consider facts as the additional contexts; that is:
\begin{equation}
\begin{aligned}
\text{Attn}&(o_{t-1}, H_C, H_F) = \Bigg[\sum_{i=1}^N \alpha^t_i h^c_i;  \sum_{j=1}^M \beta^t_j h^f_j\Bigg] \\
m^t_j &= v_f^T\text{tanh}(W^f_1 o_{t-1} + W^f_2 h^f_j) + b_f, \\
\beta^t &= \text{softmax}(m^t),
\end{aligned}
\end{equation}
and $\alpha^t$ is the same as the context-only attention. $v_f$, $W^f_1$, $W^f_2$, $b_f$ are trainable parameters.

\subsubsection{Context-Guided Fact Attention}
To better model the interaction between contexts and facts, we proposed to use the information from contexts to guide the attention towards facts.
Specifically, we modify the attention on facts from the parallel attention as below.
We first calculate the attention distribution from contexts to facts,
\begin{equation}
\begin{aligned}
M_{i,j} &= v_{g}^T\text{tanh}(W^{g}_1 h^c_i + W^{g}_2 h^f_j) + b_{g}, \\
m^c_j &= \sum_{i=1}^N M_{i, j}, \\
\beta^c &= \text{softmax}(m^c).
\end{aligned}
\end{equation}
Then, for each step, we calculate the attention from the last-step output to facts, and take the mean of two distributions as the final attention distribution on facts:
\begin{equation}
\begin{aligned}
\hat{m}^t_i &= v_o^T \text{tanh}(W^o_1 o_{t-1} + W^o_2 h^c_i) + b_o, \\
\hat{\beta}^t &= \text{softmax}(\hat{m}^t), \\
\beta^t &= \frac{\hat{\beta}^t + \beta^c}{2},
\end{aligned}
\end{equation}
where $v_g$, $v_o$, $W_1^g$, $W_2^g$, $W_1^o$, $W_2^o$, $b_g$, and $b_o$ are trainable parameters.
Hence, the obtained attention is guided by the contextual information.

\subsection{Conditional Variational Generation}
The conversations in the dataset for the DSTC7 challenge are tree-like structures, where for each context, there may be more than one reference responses.
This is also an important perspective for the natural conversations: for arbitrary dialogue contexts, there are usually not only one unique way to respond to it.

With the above consideration, we take the benefit from the variational autoencoder for this task, which has the better capability of capturing such relation than a simple seq2seq model~\cite{kingma2013auto,sohn2015learning}. 
The detail of the variational model is described below.

\subsubsection{CVAE for Dialogue Generation}
For each conversation, we represent it via four random variables: the desired response $R$, the contexts and facts, $C$ and $F$, and a latent variable $z$.
The conditional probability $p(R, z \mid C, F)$ can be rewritten as:
\begin{equation}
p(R, z \mid C, F) = p(R \mid C, F, z)p(z \mid C, F).    
\end{equation}
We model the probability $p(R \mid C, F, z)$ and $p(z \mid C, F)$ using the parameters $\theta$ and $\phi$ respectively.
Under the variational autoencoder (VAE) framework, we can interpret $\theta$ and $\phi$ as the decoder and the encoder; by setting up a Bayesian prior $p(z \mid C, F)$, our optimization target $p_\theta(R \mid C, F)$ becomes the variational lower bound (ELBO):
\begin{equation}
\begin{aligned}
\log p_\theta(R \mid C, F) \geq &-\text{KL}(q_\phi(z \mid R, C, F) \parallel p(z \mid C, F)) \\
&\ + \mathbb{E}_{q_\phi(z \mid R, C, F)}[\log p_\theta(R \mid C, F, z)].
\end{aligned}
\end{equation}
In our model, the prior $p(z \mid C, F)$ is set as $\mathcal{N}(0, I)$.

\subsubsection{Annealing Loss of KL Divergence}
As mentioned above, the optimization target, which is the variational lower bound of $\log p_\theta(R \mid C, F)$, is composed of two sub-goals: one is to minimize the KL divergence between the prior and the conditional encoder probability $q_\phi$; another is to maximize the reconstruction probability.

It is found that the model tends to minimize the KL divergence instead of reducing the reconstruction error during early training, resulting in a KL vanishing issue.
In order to alleviate the strong bias on minimization of KL divergence, we apply the annealing loss trick to scale down the effect of the KL term at the beginning of training for improving the performance~\cite{bowman2016generating}.

\subsection{Training}
The proposed model is trained to generate the responses using the CVAE objective, where the attention mechanisms enforce the responses to cover the fact-related information for \emph{knowledge-grounded response generation}.

\section{Experiments}

To evaluate the proposed model, we conduct the experiments on the DSTC7 challenge.
The used dataset and the experimental setting are described below.
Then the results are analyzed in terms of objective and subjective evaluation metrics.

\begin{table}[t!]
  \centering
  \begin{tabular}{l|c|r|r}
    \toprule
     & Time Period & Before Filter & After Filter \\
     \midrule
    Train & 2015-01$\sim$2016-12 & 1,101,684 & 142,750 \\
    Dev & 2017-01$\sim$2017-06 & 116,858 & 14,875 \\
    \bottomrule
  \end{tabular}
  \caption{Statistics of the used dataset.}
  \label{tab:data_stats}
\end{table}

\subsection{Dataset}
The dataset used in DSTC7-Track2 is crawled from Reddit with the scripts\footnote{\url{https://github.com/DSTC-MSR-NLP/DSTC7-End-to-End-Conversation-Modeling}}, which consists of discussions from subreddits like \textit{todayilearned}, \textit{worldnews}, \textit{movies}, etc.
In the dataset, the posts include a link to an external webpage, from which the facts for each conversation are then extracted.

In order to encourage our conversation model to contain the factual information, we process this dataset to make sure the conversations in which the context and provided facts are relevant.
The processing procedure is described as:
\begin{enumerate}
\item \textbf{Fact relevance}: Because the facts are extracted from the HTML source codes of webpages, some of them lack the relevant information (e.g. metadata), we use TF-IDF to rank all facts and keep the top-1 fact as the relevant knowledge for ensuring better data quality.
\item \textbf{Knowledge-grounded response}: Because the discussions in some conversations may deviate from the original topic, making all facts being irrelevant to the dialogue contexts, we thus filter out data samples where the response and the retrieved fact have no common words without considering punctuations and stopwords\footnote{We used stopwords defined in spaCy.}.
This procedure ensures the training data to match our goal about knowledge-grounded responses.
\end{enumerate}
Due to the limitation of computation resources (one GTX 1080), we use only a subset of training data, and discard the data samples with the responses longer than 20. 
Table \ref{tab:data_stats} shows the detailed statistics of the dataset after our processing.

\begin{table*}[t!]
\begin{center}
\small
\begin{tabular}{l c c c c c c c c c c c c c c c }
\toprule
\textbf{Model}      & \textbf{B-1} & \textbf{B-2} & \textbf{B-3} & \textbf{B-4} & \textbf{N-1} & \textbf{N-2} & \textbf{N-3} & \textbf{N-4} & \textbf{MET} & \textbf{Div-1} & \textbf{Div-2} & \textbf{Ent-1} & \textbf{Ent-2} & \textbf{Ent-3} & \textbf{Ent-4}       \\
\midrule
Baseline & 2.861 & .566 & .143 & .041 & .007 & .007 & .007 & .007 & 2.439 & .004 & .012 & 3.937 & 4.958 & 5.504 & 5.996          \\
\midrule
CO      & 2.634 & .513 & .130 & .038 & .006 & .006 & .006 & .006 & 2.417 & \bf .013 & .028 & 4.137 & 5.392 & 6.148 & 6.734          \\
+CVAE & 2.690 & .527 & .145 & .042 & .007 & .007 & .007 & .007 & 2.371 & \bf .013 & \bf .030 & 4.204 & \bf 5.436 & \bf 6.239 & \bf 7.049\\
\midrule
PA     & 3.698 & .763 & .200 & .063 & .020 & .021 & .021 & .021 & 2.574 & .012 & .027 & \bf 4.244 & 5.378 & 6.040 & 6.576          \\
+CVAE & 2.449 & .538 & .129 & .033 & .009 & .009 & .009 & .009 & 2.301 & .011 & .027 & 4.120 & 5.338 & 6.125 & 6.775          \\
\midrule
CG     & 2.142 & .443 & .124 & .040 & .004 & .004 & .004 & .004 & 2.258 & .011 & .026 & 4.131 & 5.300 & 6.082 & 6.820         \\
+CVAE & \bf 3.898 & \bf .817 & \bf .223 & \bf .074 & \bf .023 & \bf .024 & \bf .024 & \bf .024 & \bf 2.620 & .012 & .027 & 4.089 & 5.220 & 5.916 & 6.427          \\
\bottomrule
\end{tabular}
\end{center}
\vspace{-2mm}
\caption{The automatic evaluation results of the baselines and the proposed methods. Baseline is a context-to-response seq2seq model without attention. CO, PA, CG correspond to context-only attention, parallel attention and context-guided attention respectively. (B: BLEU; N: Nist; MET: METEOR)}
\label{tab:auto_metrics}
\end{table*}

\begin{table*}[t!]
\begin{center}
\begin{tabular}{l l c c c c c}
\toprule
& \multirow{2}{*}{\textbf{Model}} & \textbf{Context } & \multirow{2}{*}{\textbf{Interest}} & \multirow{2}{*}{\textbf{Fluency}} & \textbf{Knowledge} & \multirow{2}{*}{\textbf{Average}}\\
& & \bf Relevance & & & \bf Relatedness & \\
\midrule
\multirow{3}{*}{Offline} & PA      & 2.47$\pm$0.86 & 2.37$\pm$0.75 & 4.13$\pm$0.85 & 2.19$\pm$0.87 & 2.79 \\
& PA+CVAE & 2.40$\pm$0.81 & 2.38$\pm$0.77 & 4.00$\pm$0.92 & 2.10$\pm$0.86 & 2.72 \\
& CG      & 2.25$\pm$0.83 & 2.18$\pm$0.76 & 3.86$\pm$1.07 & 2.02$\pm$0.83 & 2.58 \\
\midrule
\multirow{3}{*}{Official} & Submitted (CG+CVAE) & 2.52$\pm$0.04 & 2.40$\pm$0.05 & - & - & 2.46 \\
& Best & 3.09$\pm$0.04 & 2.87$\pm$0.05 & - & - & 2.94\\
& Human & 3.61$\pm$0.04 & 3.49$\pm$0.04 & - & - & 3.55 \\
\bottomrule
\end{tabular}
\end{center}
\vspace{-2mm}
\caption{Human evaluation results in our offline and the official evaluation.}
\label{tab:human_metrics}
\end{table*}

\subsection{Training Details}
Considering that the dataset contains a large amount of Internet slangs and spoken English, we train a 100 dimension word embeddings via \textit{GLoVe} from train and development conversations and facts~\cite{pennington2014glove}.
We truncate the context to the last $100$ tokens and the fact to the first $500$ tokens.

The context encoder $Enc_C$ is a 2-layer bidirectional GRU \cite{cho2014learning} with hidden size 128; the fact encoder $Enc_H$ is a convolutional network with 1,2,3 width filters, and 128 feature maps per filter.
The decoder $Dec$ is a 2-layer unidirectional GRU with the hidden size 128.
For the CVAE variants, another 2-layer bidirectional GRU with the hidden size 128 is used to encode the responses.

Our models are trained using the teacher-forcing mechanism to maximize the likelihood of generating $\mathbf{R} = \{r_i\}_{i=1}^{T^r}$.
We used \texttt{adam} \cite{kingma2014adam} with the default setting as our optimizer.
During testing, we apply the beam search where the beam size is $8$.

\subsection{Results}
In the experiments, we perform two sets of evaluation, automatic evaluation and human evaluation, to better validate our generated results.

\subsubsection{Automatic Evaluation}
Our evaluation metrics include BLEU \cite{papineni2002bleu}, METEOR \cite{banerjee2005meteor}, NIST \cite{doddington2002automatic}, diversity \cite{li2016diversity} and entropy \cite{zhang2018generating} scores.
We use the implementation in the Python package nlg-eval\footnote{\url{https://github.com/Maluuba/nlg-eval}} for BLEU and METEOR scores \cite{sharma2017nlgeval}, and the NLTK toolkit to calculate NIST scores.
Our results are shown in Table \ref{tab:auto_metrics}.

It can be found that the context-guided attention model with CVAE (CG+CVAE) achieves better performance for most metrics in terms of the similarity between the generated responses and the ground truth responses.
This justifies the effectiveness of our context-guided attention, because its goal is to generate responses containing more relevant knowledge, and the metrics slightly measure the relatedness.
However, the context-only attention with CVAE (CO+VVAE) obtains the higher diversity, which is also important for this generation task.
The results show the small improvement achieved by the proposed CVAE model in terms of the generation quality and the diversity.

\subsubsection{Human Evaluation}
In order to understand the effect of our fact-grounded attention and variational generation, we conduct human evaluation on three proposed methods: the parallel attention model as our baseline (PA), compared with the parallel attention with variational generation (PA+CVAE), and the context-guided attention (CG).
First, we randomly sample 100 testing samples that fulfill the following two conditions:
\begin{enumerate}
    \item Each response has at least 3 words, because some methods tend to produce very short responses, which is hard to evaluate.
    \item Due to the goal about fact-grounded generation, we make sure that the contexts and the retrieved fact have more than 3 common words for each sample, where punctuations and stop-words are not considered.
\end{enumerate}
Then we conduct human evaluation for our proposed methods in a similar way to the official evaluation:
\begin{enumerate}
    \item In addition to \textit{relevance} and \textit{interest}, which are asked in official evaluation, we ask the judges to evaluate two additional metrics: \textit{fluency} and \textit{knowledge relatedness} (to the retrieved fact) of our response.
    \item Because we only pick one fact based on the contexts as our model input, we directly provide this fact to judges as the extra information for them to better evaluate \textit{knowledge relatedness} of the response.
\end{enumerate}
The results are shown in Table~\ref{tab:human_metrics}.
The submitted system, the best achieved results, and human performance are also included in Table~\ref{tab:human_metrics} for better comparison.
Note that the numbers for two sets of evaluation may not be directly compared but for reference.

In the offline human evaluation, it is found that the proposed models do not achieve better performance and the difference between all models are small.
From the official evaluation, our submitted results are also between \emph{disagree} (2) and \emph{neutral} (3) as in our evaluation, but the context-guided attention achieves slightly better numbers than other proposed models shown in the offline setting.
Furthermore, the best achieved performance is about 2.94, which is also lower than \emph{neutral} (3), implying the difficulty of this task.
It is clear that there is a huge gap between the currently machine-achieved and human-achieved performance, so this task requires further investigation.

\begin{table*}[t!]
\begin{center}
\begin{tabular}{p{17cm}}
\toprule
\textbf{Retrieved top-1 fact:} \\
in the united states , centenarians traditionally receive a letter from the president , congratulating them for their longevity . nbc ' s today show has also named new centenarians on air since 1983 . centenarians born in ireland receive a â‚¬ 2,540 " centenarians ' bounty " and a letter from the president of ireland , even if they are resident abroad . [ 63 ] japanese centenarians receive a silver cup and a certificate from the prime minister of japan upon their 100th birthday , honouring them for their longevity and prosperity in their lives . swedish centenarians receive a telegram from the king and queen of sweden . [ 64 ] centenarians born in italy receive a letter from the president of italy . in japan , a " national respect for the aged day " has been celebrated every september since 1966 . \\
\midrule
\textbf{Conversation:} \\
-- til in the united states , people who turn 100 years old receive a letter from the president , congratulating them on their \\
\hspace{0.2cm} longevity . \\
-- same in canada but 90 instead of 100 \\
\midrule
\textbf{Ground Truth:} is that the canadian exchange rate these days ? \\
\textbf{PA Response:} they are the same thing . \\
\textbf{PA+CVAE Response:} you can have to be a . \\
\textbf{CG Response:} it's not the same thing in the uk . \\
\bottomrule
\end{tabular}
\end{center}
\caption{Model response sample.}
\label{tab:model_response_sample}
\end{table*}

\subsection{Qualitative Analysis}
The above results tell that there is no significant difference between our proposed models and baselines.
A sample model responses from the human evaluation set is shown in Table~\ref{tab:model_response_sample} for our qualitative analysis.
In this example, adding CVAE generates a more diverse response than the parallel attention result, but may not effectively ground the knowledge in the sentence.
Also, our context-guided result seems to focus more on the fact compared to other models.
However, the ground truth in the data is very difficult to simulate for the current models, because it may need additional knowledge or common sense.
From the current results achieved by our model, we conclude that this task still needs further investigation.

\section{Conclusion}
We describe a variational knowledge-grounded conversation system, which attempts at modeling the relations between dialogue contexts and external facts in an end-to-end fashion.
It guides a potential research direction about how external information interacts with dialogues and how the machine can capture such interaction for better knowledge-grounded response generation.
In the experiments on DSTC7, the results demonstrate the difficulty of this task, because almost all current models fail to generate reasonable responses.
Therefore, the knowledge-grounded dialogue modeling requires further study in order to advance the machine's capacity of producing a informative and knowledgable conversation.


\end{document}